\pdfoutput=1

\documentclass[11pt]{article}

\usepackage[]{acl}

\usepackage{times}
\usepackage{helvet}
\usepackage{courier}
\usepackage{booktabs}
\usepackage{graphicx}
\usepackage[labelformat=simple]{subcaption}

\usepackage{amsmath}
\usepackage{xcolor}
\usepackage{latexsym}
\usepackage{tikz}
\usepackage{etoolbox}
\usepackage{booktabs}
\usepackage{tikzsymbols}
\usepackage{adjustbox}
\usepackage{verbatim}
\usepackage{caption}
\usepackage{multirow}
\usepackage{soul}
\usepackage{natbib}
\usepackage{amsfonts}
\usepackage{float}
\usepackage{amssymb}
\usepackage{rotating}

\UseRawInputEncoding
\usepackage[T1]{fontenc}

\usepackage[utf8]{inputenc}

\usepackage{microtype}

\title{A Survey on Recent Advances and Challenges in Reinforcement Learning Methods for Task-Oriented Dialogue Policy Learning}

\author{Wai-Chung Kwan$^*$, Hongru Wang$^*$, Huimin Wang \and Kam-Fai Wong   \\
  The Chinese University of Hong Kong\\
  \texttt{\{wckwan, hrwang, hmwang, kfwong\}@se.cuhk.edu.hk} \\}

\begin{document}
\maketitle
\def\thefootnote{*}\footnotetext{These authors contributed equally to this work.}\def\thefootnote{\arabic{footnote}}
\begin{abstract}
Dialogue Policy Learning is a key component in a task-oriented dialogue system (TDS) that decides the next action of the system given the dialogue state at each turn. Reinforcement Learning (RL) is commonly chosen to learn the dialogue policy, regarding the user as the environment and the system as the agent. Many benchmark datasets and algorithms have been created to facilitate the development and evaluation of dialogue policy based on RL. In this paper, we survey recent advances and challenges in dialogue policy from the prescriptive of RL. More specifically, we identify the major problems and summarize corresponding solutions for RL-based dialogue policy learning. Besides, we provide a comprehensive survey of applying RL to dialogue policy learning by categorizing recent methods into basic elements in RL. We believe this survey can shed a light on future research in dialogue management.
\end{abstract}

\section{Introduction} \label{introduction}

\begin{figure}[t]
    \centering
    \includegraphics[width=1.0\columnwidth]{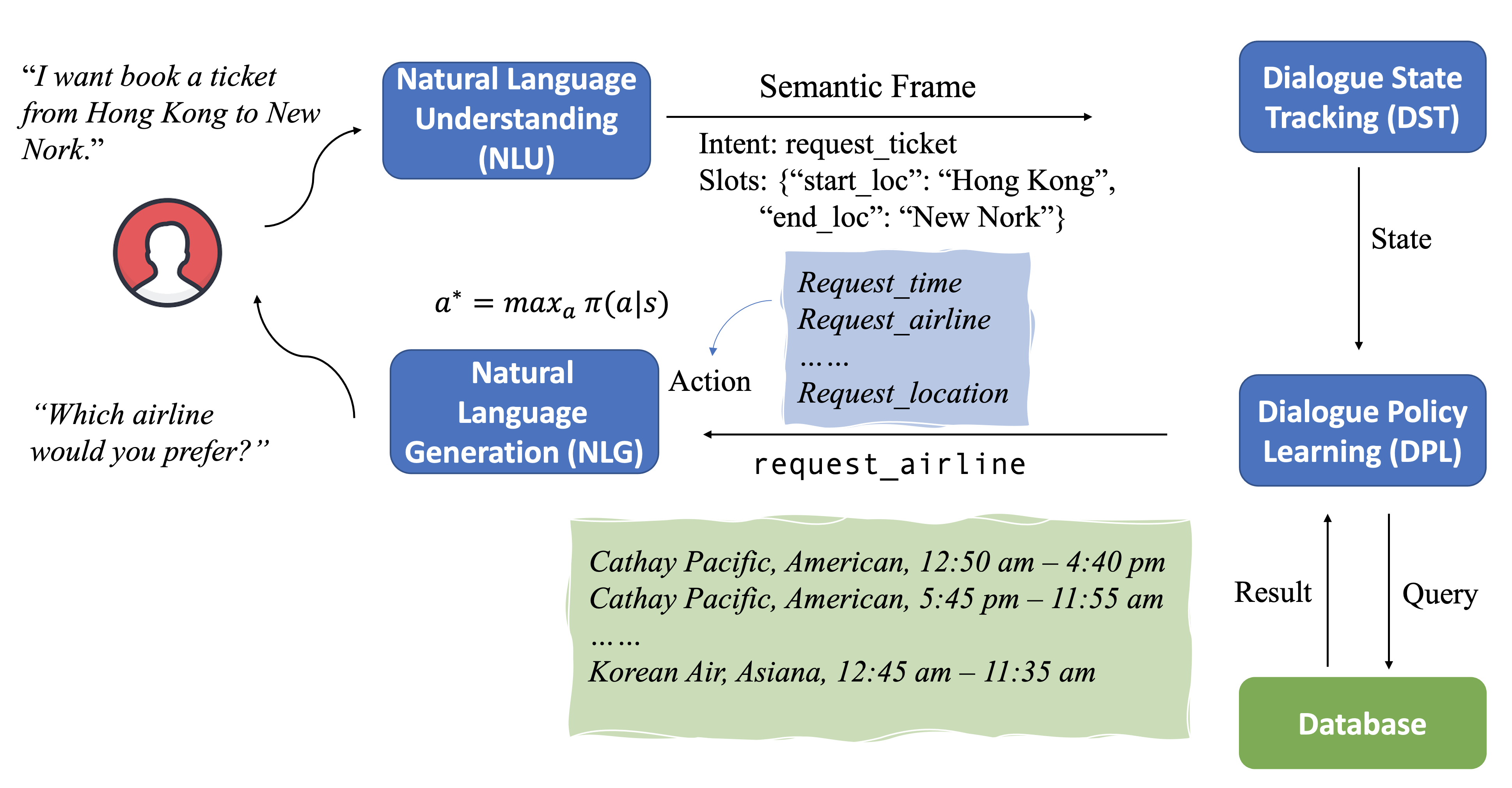}
    \caption{ A overview of task-oriented dialogue system. All \textcolor{blue}{blue} parts represent the four components in pipeline dialogue system.
	}\vspace*{-4mm}
    \label{fig:overview}
\end{figure}

TDS aims to assist users to accomplish tasks ranging from weather inquiry to schedule planning \citep{chen2017survey}. The architecture of TDS can be classified into two classes. The first one is an end-to-end approach that directly maps the user's utterance to the system's natural language response \citep{lewis-etal-2017-deal,eric2017copy,chi_speaker_2017,wang2020multi}. These works often adopt a sequence-to-sequence model and train in a supervised manner.  The second one is a pipeline approach that separates the system into four interdependent components: natural language understanding (NLU), dialogue state tracking (DST), dialogue policy learning (DPL) and natural language generation (NLG) as shown in Figure~\ref{fig:overview} \citep{liu_adversarial_2018,chen2019bert,wu2019transferable,li-etal-2020-slot}. 

Under this pipeline approach, the NLU module first recognizes the intents and slots from the input sentence given by a human. Then, the DST module represents it as an internal dialogue state. Next, the DPL module performs an action to satisfy the user. Finally, The NLG module transforms the action transformed into natural language. The end-to-end approach is more flexible and has fewer requirements on the data annotations formats. However, it requires a large amount of data and its black box structure gives us no interpretation and controls \citep{gao_dialog_2019}. The pipeline approach is more interpretable and easier to implement. Although the whole system is harder to optimize globally, it's preferred by most commercial dialogue systems \citep{zhang_recent_2020}. In the pipeline approach, DPL plays a key role in TDS as an intermediate juncture between the DST and NLG components.

In recent years, we have witnessed the prosperity of the application of RL in DPL. \citet{levin_learning_1997} is the first work that treats DPL as an MDP problem. It outlines the complexities of modelling DPL as an MDP problem and justifies the use of RL algorithms to optimize the MDP problem. Thereafter, there exist few works that extend the RL approach and identify the challenge in approximating the dialogue state \citep{walker2000application,singh_reinforcement_2000,singh2002optimizing}. 

At the other end of the spectrum, several researchers explored using supervised learning (SL) techniques in DPL \citep{gandhe_creating_2007,henderson_hybrid_2008,devault_toward_2011,vinyals_neural_2015,shang_neural_2015}. The main idea was to train the model to output the next system action given the dialogue context. However, SL does not consider the future effects of the current decision which may lead to sub-optimal behaviour \citep{henderson_hybrid_2008}.

With the breakthroughs in deep learning, deep reinforcement learning (DRL) methods that combine neural networks with RL has recently led to successes in learning policies for a wide range of sequential decision-making problems. This includes simulated environments like the Atari games \citep{mnih_playing_2013}), the chess game Go \citep{silver_mastering_2016} and various robotic tasks \citep{ng_autonomous_2006,peters_reinforcement_2008}. Following that, DRL have been receiving a lot of attention and achieved successful results mainly in single domain dialogue scenario \citep{su_continuously_2016,fatemi_policy_2016,su_sample-efficient_2017,lipton_bbq-networks_2017}. The neural models can extract high-level dialogues states that encode the complicated and long language utterances, which is the biggest challenge that the early works were facing \citep{levin_learning_1997,singh_reinforcement_2000}. As the focus of DPL research has slowly gravitated to more complicated multi-domain datasets, many RL algorithms face scalability problems \citep{cuayahuitl_deep_2016}.

Recently, there has been a flurry of works that focus on ways to adapt and improve RL agents in the multi-domain scenario. Few works attempt to review the vast literature on the recent application of RL in DPL of TDS.  \citet{grasl_survey_2019} surveyed the use of RL in the four types of the dialogue system, namely social chatbots, infobots, task-oriented and personal assistant bots. However, the progress and challenges of using RL in TDS were not discussed.  \citet{dai_survey_2020} reviewed the recent progress and challenges of dialogue management which only contains a limited discussion on RL methods in DPL due to its wide scope of interest. While they pointed out the three main shortcomings of dialogue management that recent works have been addressing, a taxonomy of the methodologies is not provided. A comprehensive survey that summarizes the recent challenges and methodologies of applying RL in DPL of TDS is still lacking which motivates this survey.

In this survey, we will focus our discussion on three main recent challenges of applying RL to DPL of TDS, namely \textit{exploration efficiency}, \textit{cold start problem} and \textit{large state-action space}. These are the prominent challenges in recent work on DPL that the majority of recent literature are trying to address.  The procedure that we use to shortlist the papers for review is provided in Appendix \ref{appendix:table}. We will also give an overview of recent works that tackle those challenges. The remainder of this paper is organized as follows. In Section \ref{overview}, we first provide the problem definition of DPL and elaborate on the recent challenges of using RL to train a dialogue agent in TDS. Then, we motivate and introduce one of the contributions of this survey, a typology of recent DPL works that tackle the mentioned challenges in DPL. The typology is based on the five elements in RL: \textit{Environment}, \textit{Policy}, \textit{State}, \textit{Action} and \textit{Reward}. which are discussed separately in Section \ref{sec:environment}, \ref{sec:policy}, \ref{sec:state space}, \ref{sec:action space}, \ref{sec:reward} respectively. The topology is motivated by the fact that the key differentiating aspect of recently proposed methods can be boiled down to these five fundamental elements of RL. This allows us to highlight the similarities and differences between the methods.

In Section \ref{future direction}, we present the challenges of applying RL dialogue agents in real-life scenarios and three promising future research directions. Finally, we conclude the survey in Section \ref{conclusion}.

To sum up, the contributions are:
\begin{itemize}
    \item We identify the three recent challenges of applying RL to DPL of TDS. 
    \item We propose a general typology that characterizes the main research directions to tackle the challenges and provide a compact overview of them. 
    \item We outline the outstanding challenges in DPL of TDS and identify three fruitful future directions.
\end{itemize}

\section{Overview} \label{overview}

\subsection{Problem Definition and Annotations}
\begin{figure}[h]
    \centering
    \includegraphics[width=1.0\columnwidth]{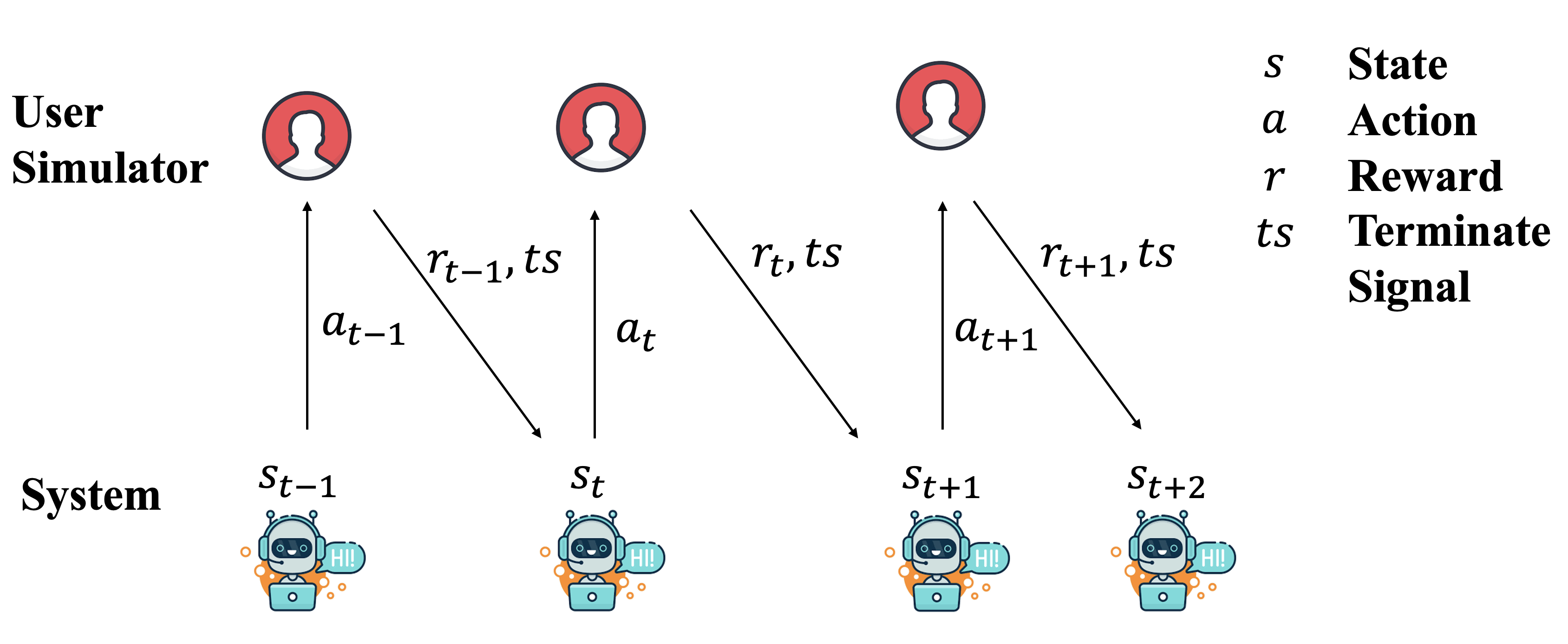}
	\caption{ The framework of Markov Decision Process in DPL. At time $t$, the system takes an action $a_t$, receiving a reward $r_t$ and a terminate signal $t$ and then transferring to a new state $s_{t+1}$.
	}\vspace*{-4mm}
    \label{fig:mdp_framework}
\end{figure}

The dialogue policy is responsible to generate the appropriate next system action given the dialogue state. DPL is often formulated as a MDP problem and RL is often used to optimize the policy \citep{liu_iterative_2017,peng_adversarial_2018,peng_deep_2018,liu_adversarial_2018,zhang_recent_2020,gordon-hall_show_2020,cao_adaptive_2020}. 
Formally, an MDP is defined as a five element tuple (S, A, P, R, $\gamma$). $\mathcal{S}$ refers to the dialogue state space that holds the necessary information for the policy to make a decision. $\mathcal{A}$ refers to the set of all system actions.  $P(s^\prime|s,a)$ refers to the transition model $S \times A \times S \rightarrow [0,1]$ of the environment. $R(s, a)$ is the reward function $S \times A \rightarrow \mathbb{R}$ that provides an immediate reward for each turn. $\gamma \in (0, 1]$ is the discount factor. Figure \ref{fig:mdp_framework} provides an overview of the MDP framework.

A full turn of dialogue interactions can be viewed as a trajectory ($s_1, a_1, r_1, \ldots)$ which is generated by the following process in each step. First, the dialogue agent observes the current environment states $s_t \in \mathcal{S}$ and responds with an action $a_t \in \mathcal{A}$. Second, the environment receives the action and transits to a new state $s_{t+1} \in \mathcal{S}$ according to the transition model. Third, the environment emits a reward $r_t$ after transiting to a new state. 

At each step t, this process gives us a tuple ($s_t, a_t, r_t, s_{t+1}$) which is called a transition. The goal of the RL agent is to learn an optimal deterministic policy $\pi : S \rightarrow A$ that maximizes the value function which is the expected total discounted returns in a trajectory. It is formally defined as $$V^\pi(s) := \mathbb{E}\left[\sum_{t=0}^T \gamma^t r_t|s_0=s\right]$$

\subsection{Recent Challenges in Applying RL}
In recent years, DPL researches aim at tackling three main challenges in using RL to train a dialogue agent in a TDS.

\textbf{1. Exploration Efficiency. }
It is arduous to find good data sources for an RL agent to learn.
RL interacts with an environment to collect interactions for training. In the dialogue system setting, the agent is required to interact with real users \citep{su_-line_2016} which is expensive and time-consuming. In practice, the agent interacts with a rule-based user simulator \citep{schatzmann_agenda-based_2007,su_continuously_2016}. The exploration efficiency depends on how closely the simulator resembles human behaviour, which is not easy \citep{walker_paradise_1997,liu_iterative_2017}. It is laborious to build high quality and specialized user simulator for a dataset.

\textbf{2. Cold Start Problem. }
A poorly initialized policy may lead to low-quality interactions with users in online learning settings \citep{chen_-line_2017}. Having rare successful experiences causes the model to learn slowly in the beginning and discourages real users to interact with the system \citep{lu_goal-oriented_2018,lu_autoeg_2020}.

\textbf{3. Large State-Action Space. }
DPL for some complex dialogue tasks such as multi-domain involve a large state-action space \citep{peng_adversarial_2018,gordon-hall_learning_2020}. The dialogue agent is required to explore in this large space and often takes many conversation turns to fulfil a task. The long trajectory results in a delayed and sparse reward, which is usually provided at the end of a conversation \citep{liu_adversarial_2018}.

\subsection{Typology of Approaches}
RL system is composed of five elements: \textit{environment}, \textit{policy}, \textit{state}, \textit{action} and \textit{reward}. All the proposed approaches that improve dialogue RL agent can be boiled down to modifications to those five elements. This motivates us to classify the recent approaches in RL dialogue agents by these five elements.  This typology not only enables us to outline similarities and differences between different approaches in a concise manner, but it also allows us to identify the focal points of recent advancement of RL methods in DPL researches starkly.

\begin{figure}[t]
    \centering
    \includegraphics[width=.5\textwidth]{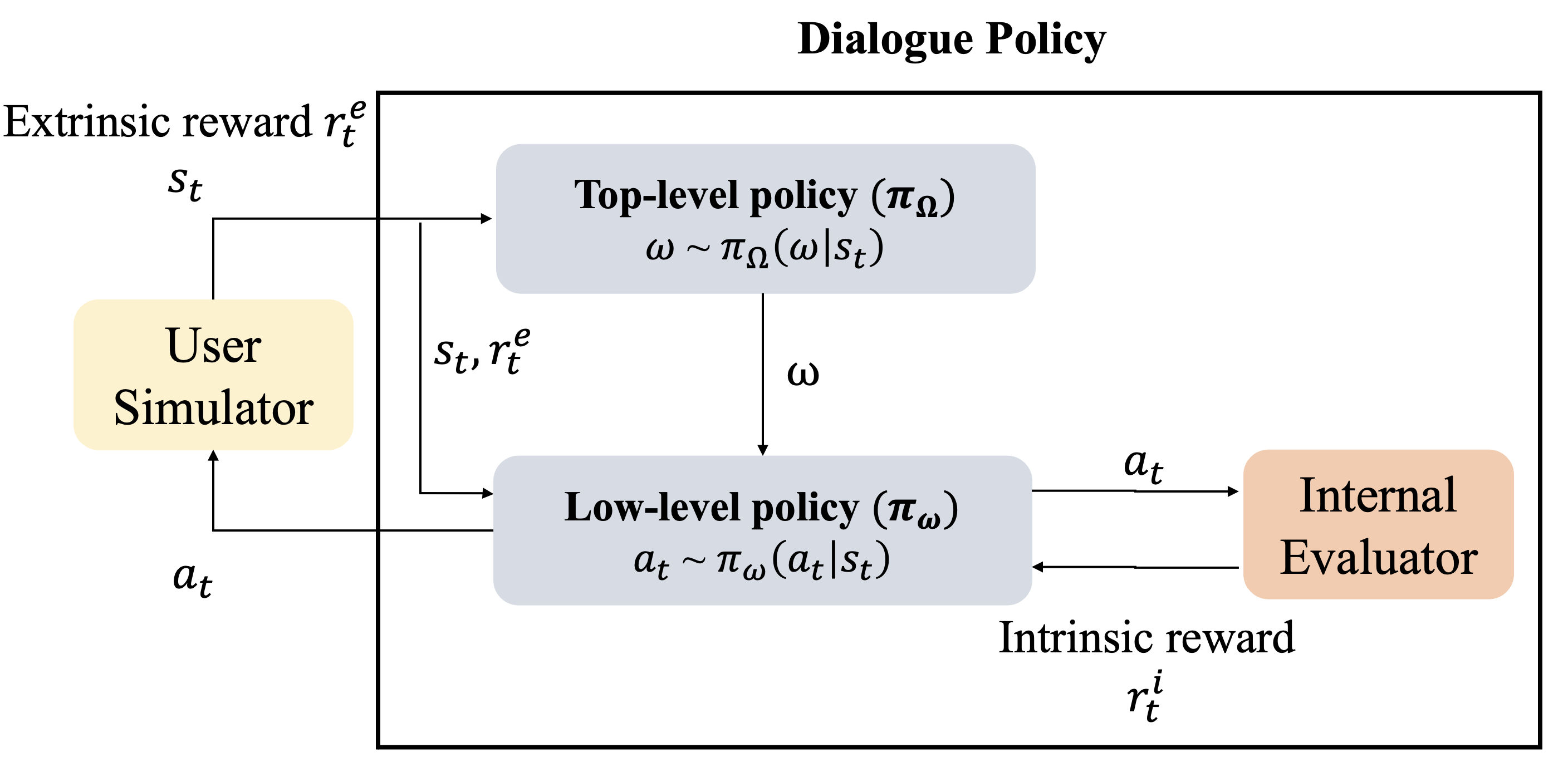}
    \caption{The overview of two levels of policies in hierarchical reinforcement learning, \citet{peng_composite_2017}}
    \label{fig:hrl}
\end{figure}

\section{Environment} \label{sec:environment}
In a typical scenario of DPL, there are two speaker roles: user and system. Most of the current methods are single-agent that only model the system side, regarding the user side as the environment \citep{su_learning_2015, su_reward_2015, su_continuously_2016, peng_composite_2017,su_sample-efficient_2017, gordon-hall_show_2020,li_guided_2020}. Few methods model two roles in $n$ dialogues \citep{liu_iterative_2017,papangelis-etal-2019-collaborative, zhang_learning_2020} and rare works consider the multi-person (more than two persons) dialogue . In this section, we will illustrate (1) different methods to build a user simulator (i.e. the environment), and (2) how to model different agents simultaneously.

\subsection{Single-Agent / User Simulator} 
Most previous works build a user simulator first and interact with the single system agent using the simulator to obtain a large number of simulated user experiences for RL algorithms. Building a reliable user simulator, however, is not trivial and often requires much expert knowledge or abundant annotated data \citep{takanobu_multi-agent_2020}. There are two major methods to build a user simulator.

\noindent \textbf{Agenda-based simulator:} With the growing need for the dialogue system to handle more complex tasks, it will be more challenging and laborious to build a fully rule-based user simulator, which requires extensive domain knowledge and expertise. An Agenda-based simulator \citep{schatzmann_agenda-based_2007,schatzmann2009hidden,li2016user,ultes2017pydial} starts a conversation with a randomly generated user goal that is unknown to the dialogue manager. It keeps a stack data structure (i.e. \textit{user agenda}) during the course of the conversation. Each entry in the stack maps to an intention the user aims to achieve, and the order follows the first-in-last-out operations of the agenda stack \citep{gao2018neural}. An agenda-based simulator stores all information that the user needs to inform and acquire, acting according to pre-defined rules.

\noindent \textbf{Data-driven simulator:} Another method to build a user simulator is to utilize a sequence-to-sequence framework, aiming to generate user response (utterance or dialogue actions) given current dialogue context \citep{sutskever2014sequence}. The dialogue context consists of historical dialogue content, dialogue goal, constraint status and request status. This method can be learned and optimized directly from a large amount of human-human dialogue corpora \citep{eckert1997user,levin2000stochastic,chandramohan2011user,asri2016sequence}.

Although there are several ways to build a user simulator, the gap between user simulator and humans make the dialogue policy optimization harder \citep{gao2018neural}. Besides, it remains challenging to evaluate the quality of a user simulator, as it is unclear to define how closely the simulator resembles real user behaviours \citep{williams2008evaluating,ai2008assessing,pietquin2013survey}.

\subsection{Multi-Agents}  \label{sec:multi-agent}
The goal of RL is to discover the optimal strategy $\pi^* (a|s)$ of the MDP. It can be extended into the N-agents setting where each agent has its own set of states $S_i$ and actions $A_i$. In Multi-Agent Reinforcement Learning (MARL), the state transition $s = (s_1, . . . , s_N) \longrightarrow s^{'} = (s_1^{'} , . . . , s_N^{'})$ depends on the actions taken by all agents $(a_1 , . . . , a_N )$ according to each agent’s policy $\pi_i (a_i | s_i)$ where $s_i \in S_i, a_i \in A_i$, and similar to single-agent RL, each agent aims to maximize its local total discounted return $R_i = \sum_t \gamma^{t}r_{i,t}$. 

Instead of employing a user simulator, \citet{georgila-etal-2014-single} demonstrated that two agents learn concurrently by interacting with each other
without any need for simulated users can achieve satisfactory performance in a negotiation scenario. \citet{liu_iterative_2017} makes the first attempt to apply MARL into the task-oriented dialogue policy to learn the system policy and user policy concurrently. It optimizes two agents from the corpus by iteratively training the system policy and the user policy with the policy gradient method. Thereafter,  \citet{papangelis-etal-2019-collaborative} applied WoLF-PHC within the MARL framework into the task-oriented dialogue policy, which is based on Q-learning for mixed policies to achieve faster learning. Following this line of research, \citet{takanobu_multi-agent_2020} scaled it to multi-domain dialogue by using the actor-critic framework instead to deal with the large discrete action space in dialogue. Recent work extends traditional two-agent to three-agent, leading to smaller action space and faster learning \citep{wang-wong-2021-collaborative}. Another work explores the MARL framework in a different perspective \citep{gavsic2015multi}. They use MARL in the policy committee framework where each policy decides an action on its own and is combined by a gating mechanism.

\section{Policy} \label{sec:policy}
In this section, we firstly divide different DPL methods into two categories: \textit{model-free reinforcement learning} and \textit{model-based reinforcement learning}, in which the former methods can be further divided into \textit{hierarchical reinforcement learning} (i.e, HRL) \citep{NIPS1997_5ca3e9b1,dietterich2000hierarchical}  and \textit{feudal reinforcement learning} (i.e, FRL) \citep{6407655}. In addition, most of these methods requires warm up before training which is illustrated at the last.

\begin{figure}[t]
    \centering
    \includegraphics[width=.3\textwidth]{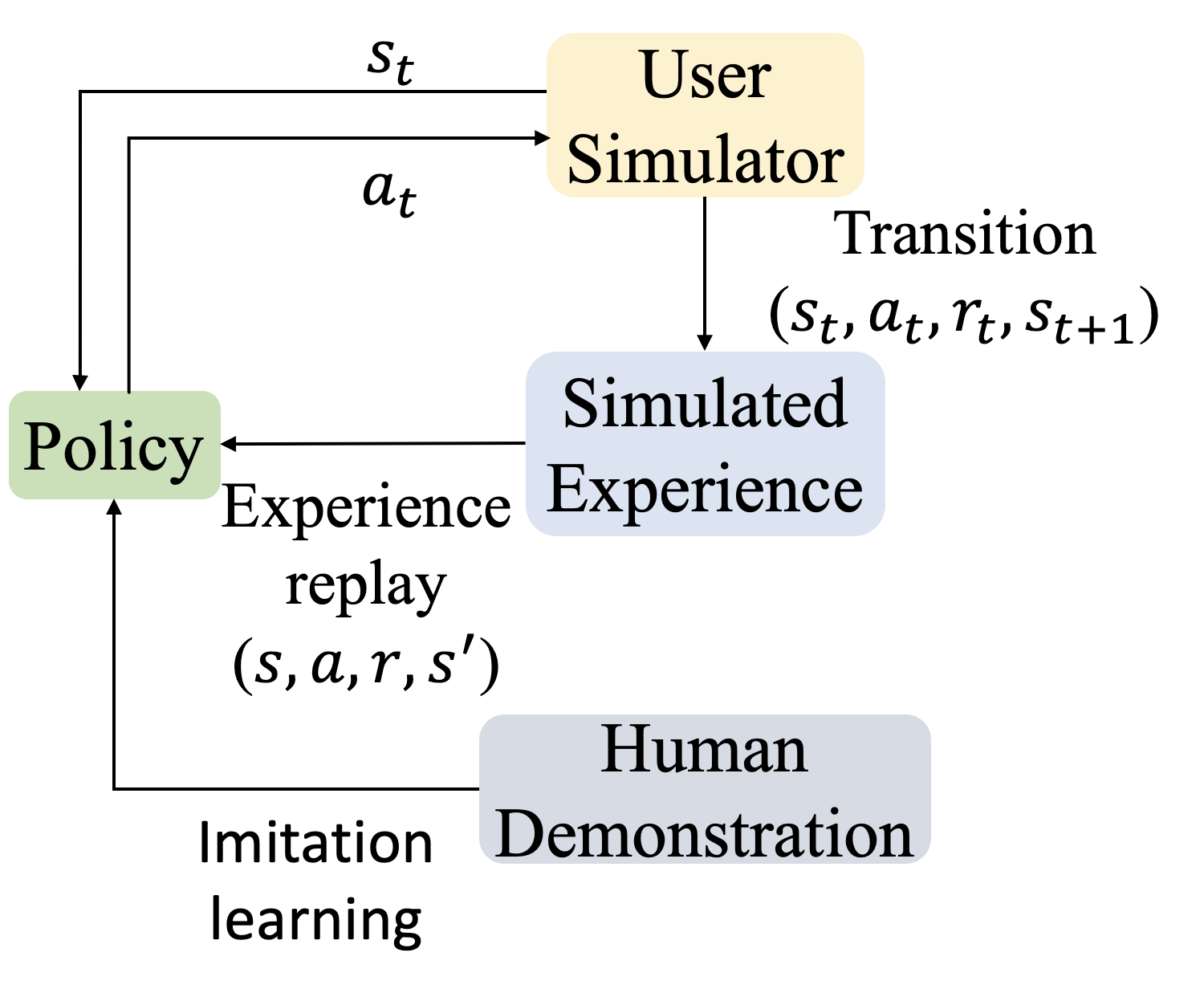}
    \caption{The RL architecture of using imitation learning.}
    \label{fig:imitation learning}
\end{figure}

\subsection{Model-Free RL - HRL}
Solving composite tasks, which consist of several inherent sub-tasks, remains a challenge in the research area of dialogue systems. For instance, a composite dialogue of making a hotel reservation involves several sub-tasks, such as looking for a hotel that meets the user’s constraints, booking the room and paying for the room. HRL decomposes complex tasks into several subtasks and learns different policies for these subtasks from top to low-level \citep{budzianowski-etal-2017-sub,peng_composite_2017,kristianto-etal-2018-autonomous}. As shown in figure \ref{fig:hrl}, the top-level policy decides which option (i.e. subtask) $w \in \varOmega$ should be chosen, and the low-level dialogue policy selects the primitive actions $a \in A$ to complete the subtask given by the top-level policy. It is noted that a primitive action is an action lasting for one time step, while an option is an action lasting for several time steps. According to the realm of the top-level policy, HRL can be further divided into sub-domain or sub-goal hierarchical reinforcement learning. 

\textbf{Sub-domain. } \citet{peng_composite_2017,budzianowski-etal-2017-sub} used the options framework \citep{sutton_between_1999} to solve the above problem with different approximators. However, each option (i.e. sub-task) and its property (e.g. starting and terminating conditions, and valid action set) had to be manually defined in their works. \citet{kristianto-etal-2018-autonomous} proposed a unified framework that integrates option discovery \citep{bacon_option-critic_2016,machado_laplacian_2017} and achieved a comparable performance with manually defined options framework.
    
\textbf{Sub-goal. } Instead of decomposing a task according to the corresponding domain, it is also an option to divide a complex goal-oriented task into a set of simpler subgoals. \citet{tang_subgoal_2018} proposed the Subgoal Discovery Network (SDN) that discovers and exploits the hidden structure of the task to enable efficient policy learning inspired by the sequence segmentation model \citep{wang2017sequence}.

\begin{figure*}[t]
  \centering
  \subcaptionbox{Inverse reinforcement learning. \label{fig:inverse reinforcement learning}}[.45\textwidth][c]{
    \includegraphics[width=.45\textwidth]{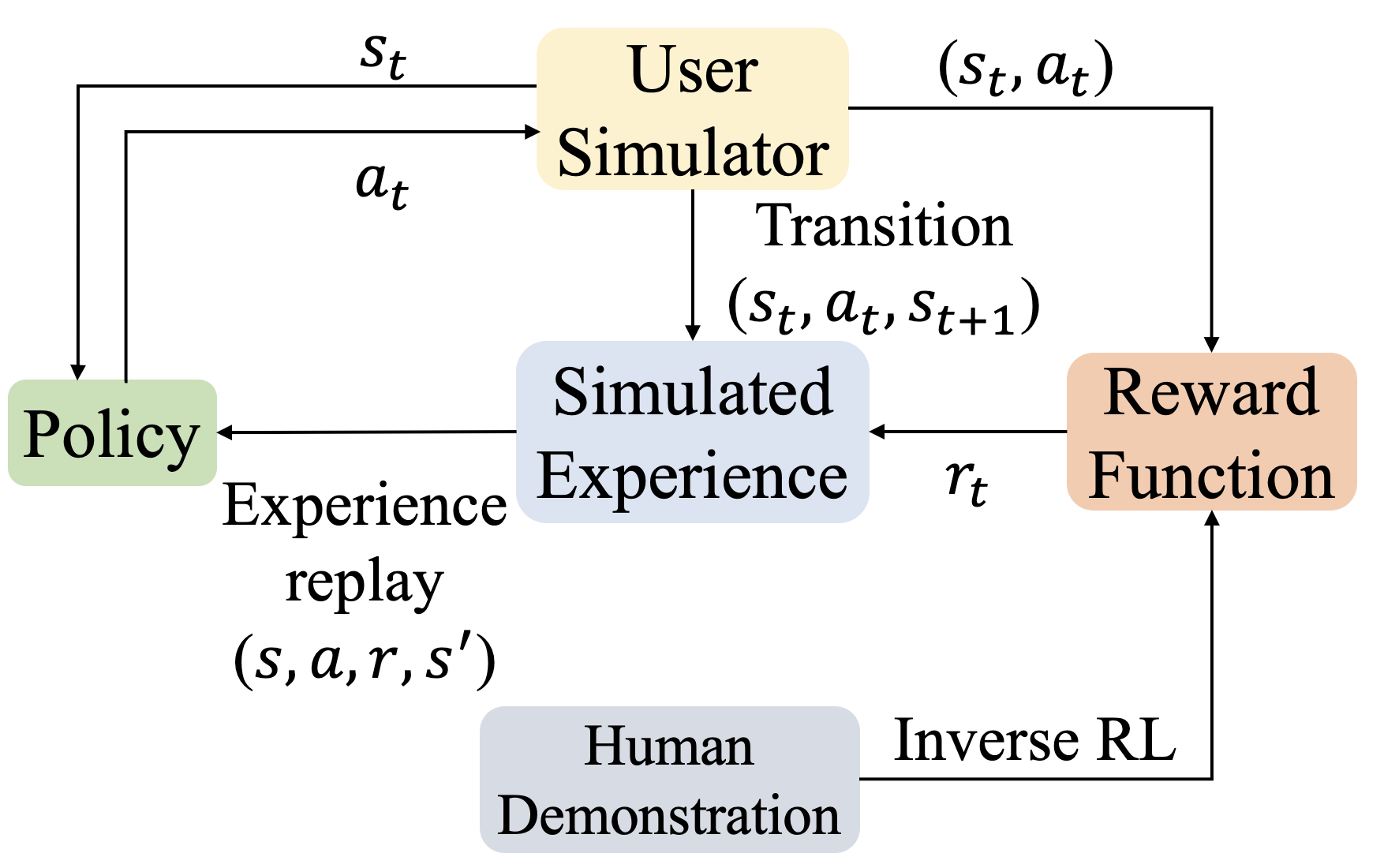}}\quad
  \subcaptionbox{Reward shaping. \label{fig:reward shaping}} [.45\textwidth][c]{
    \includegraphics[width=.45\textwidth]{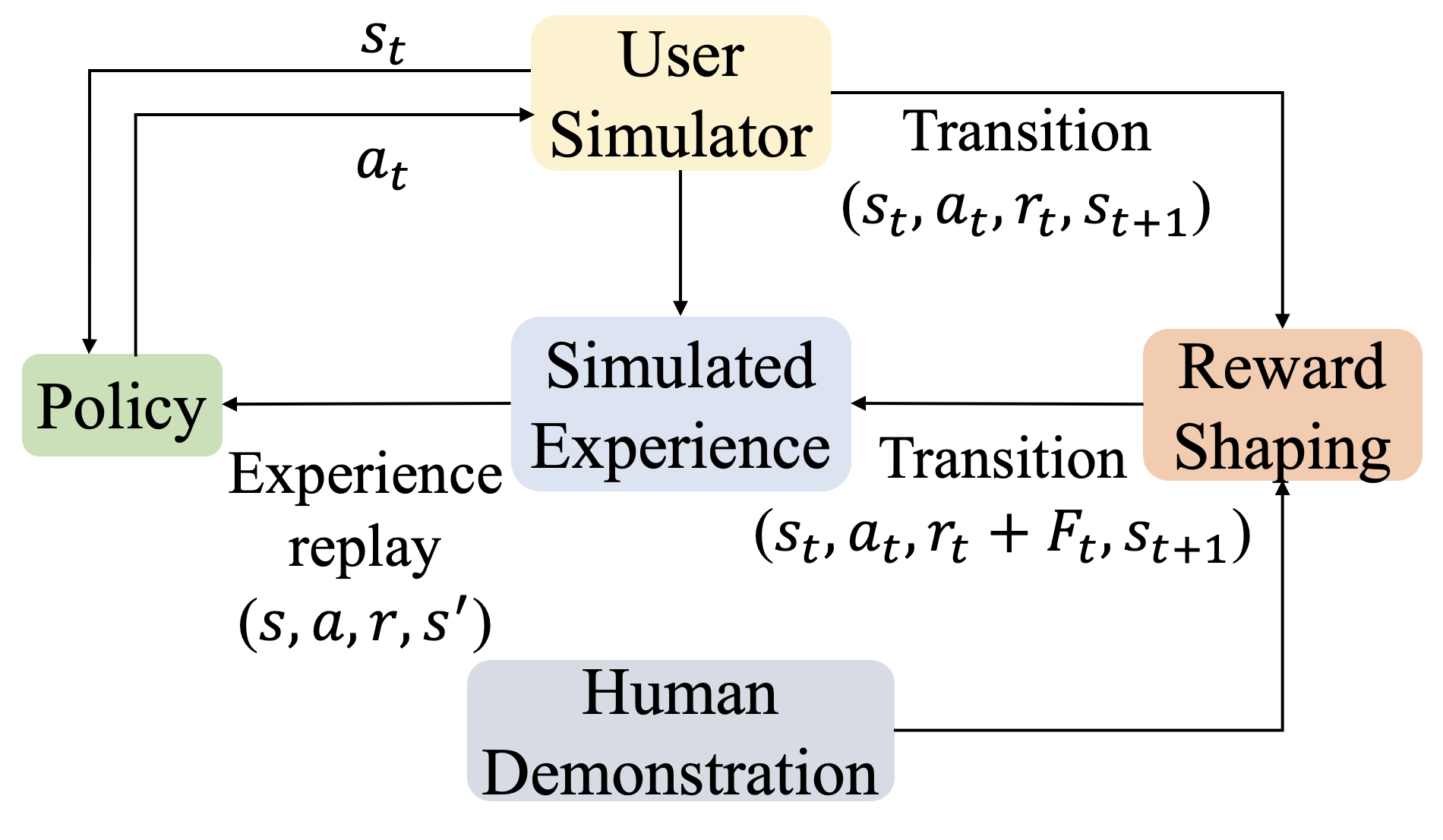}}
   
    \caption{Two strategies to learn a denser reward. \label{fig:reward learning}}
\end{figure*}

\subsection{Model-Free RL - FRL}

Feudal Reinforcement Learning (FRL) \citep{10.5555/645753.668239} is another interesting attempt to solve the large state and action space problem. FRL decomposes a task \textit{spatially} to restrict the action space of each sub-policy, but the above mentioned HRL decompose a task \textit{temporarily} to solve a different sub-task at a different time step \citep{gao2018neural,dai_survey_2020}. \cite{casanueva2018feudal} firstly applied FRL to task-oriented dialogue systems and decomposes the decision into two steps based on its relevance with slots: a master policy is chosen to select a subset of primitive actions at the first step, and a primitive action is chosen from the selected subset at the second step. The decisions in different steps use different parts of the abstracted states. Furthermore, \cite{casanueva-etal-2018-feudal-dialogue} showed that the feature extraction can be learned jointly with the policy model while obtaining similar performance, even outperforming the handcrafted features in feudal dialogue management.

In contrast to the HRL that decompose a task into \textit{temporally} separated subtasks, FRL decomposes a complex decision \textit{spatially} \citep{gao2018neural}. Although both HRL and FRL can be used to address large dimension issues, they both have their notorious limitation: the decomposition in HRL often requires expert knowledge while FRL does not consider the mutual constraints between sub-tasks \citep{dai_survey_2020}.

\subsection{Model-Based RL} \label{sec:model-based}
Different from model-free RL methods, model-based RL models the environment to decide the transition of states, enabling planning for dialogue policy learning \citep{zhang_recent_2020}. Deep Dyna-Q (DDQ) \citep{peng_deep_2018} is the first deep RL framework that integrates planning for task-completion DPL, which effectively leverages a small number of real conversations. Specifically, the environment is modelled as a \textit{world model} to mimic the real user response and generate simulated experience. Recently, more DDQ variants have been proposed to improve the quality of simulated experience by adversarial training \citep{su_discriminative_2018}, active learning \citep{wu_switch-based_2018} and human teaching \citep{zhang_budgeted_2019}. 

\subsection{Warm-up by Imitation Learning} \label{sec:il}

Imitation Learning (IL) allows the policy to imitate directly from the expert demonstrations without exploring the environment, leading to an effective initialization at the warm-up stage \citep{abbeel_apprenticeship_2004}. With limited warm-up steps based on a few expert demonstrations, the learning speed of the dialogue RL agent can be accelerated \citep{su_continuously_2016,fatemi_policy_2016}. However, another line of works points out that IL requires expert demonstrations and the transition dynamics of the RL environment to have the same distribution, which is often not the case in DPL. Thus, it's critical to follow up the IL with different RL methods \citep{liu_iterative_2017,peng_deep_2018}. 


\section{State Space} \label{sec:state space}
The dialogue state encodes the essential information in the dialogue history for the dialogue policy to generate the next system action. There are mainly two types of states representation that were used by recent researches. They are the multi-hot representation and the distributed representation.

Most works using the multi-hot representation are based on a belief vector that simply concatenates the one-hot vector for each slot \citep{takanobu_guided_2019,takanobu_multi-agent_2020, xu_meta_2020,jhunjhunwala_multi-action_2020}. These multi-hot representations are often simple to implement but require features engineering.   On the other hand, some works \citep{liu_iterative_2017,wu_switch-based_2018,peng_deep_2018} adopted the approach in \citet{mrksic_neural_2017} where the state representations were directly learned from user's utterances. \citet{saha_towards_2020} extended the state representation with multi-modal information. They added image and sentiment representations into the state. This approach requires no human intervention and enables to handle variations \citep{mrksic_neural_2017}.

\section{Action Space} \label{sec:action space}
Most works treat the action space as the set of dialogue acts. A dialogue act is specified by a dialogue act type which indicates the type of action the user/agent is performing, and a set of slot-value pairs that specify the imposed constraints \cite{de_mori_spoken_2007}. 

\citet{chen_semantically_2019} pointed out that having a separate set of dialogue acts for each domain is not scalable as we work towards multi-domain large-scale scenarios. They alleviated this problem by building a multi-layer hierarchical graph to exploit the structure of dialogue acts. While this work has avoided the dialogue acts to grow exponential with the number of domains, \citet{zhao_rethinking_2019} took another approach to treat the action space as a latent variable and use an unsupervised method to induce an appropriate action space from the data. 

At the other end of the spectrum, some works represent dialogue acts as sequences and formulate the dialogue act prediction problem as a sequence generation problem \citep{shu2019modeling}. The advantage of this method is its ability to output multiple actions per turn. Most existing methods for DPL that are formulated as a classification problem can only predict one system action per turn.

\section{Reward Learning} \label{sec:reward}

Most works adopted the manually designed reward function that gives large positive and negative reward for success and failed dialogue respectively and a small negative turn level reward to encourage shorter dialogue \cite{asri_task_2014,su_learning_2015,su_-line_2016,fatemi_policy_2016,su_sample-efficient_2017,peng_composite_2017,peng_deep_2018,lu_goal-oriented_2018,kristianto-etal-2018-autonomous,su_discriminative_2018,tang_subgoal_2018,weisz_sample_2018,wu_switch-based_2018}. However, the sparse rewarding signal of is one of the reasons that RL agents have poor learning efficiency \cite{takanobu_guided_2019,wang_learning_2020}.

Below, we present two streams of work that aim to learn a denser reward to encourage faster learning in RL making using of the provided expert demonstrations: inverse reinforcement learning IRL based methods and reward shaping. Figure~\ref{fig:reward_learning} shows the overview of the pipeline of IRL methods and reward shaping.

\subsection{Inverse Reinforcement Learning Method} \label{sec:irl}
IRL is a fundamental technique to learn a reward function that underlies the expert demonstrations \citep{russell_learning_1998}. \citet{boularias_learning_2010} is the first to explore this idea in DPL to learn a reward function from a human expert in a Wizard-of-Oz setting. The proposed a reward function which is a linear combination of feature vectors with unknown weights. The weights can be first learned from the expert demonstrations, then the learned reward function is used in RL. The learned reward function can provide meaningful feedback to the policy which helps it to learn effectively especially in the early stage.

IRL is often expensive to run which hinders it to scale to a more complex dialogue scenario \cite{ho_generative_2016}. In the RL community, Adversarial IRL (AL-IRL) is proposed to enhance the learning efficiency to learn the reward from expert demonstrations \cite{ho_generative_2016}. \citet{liu_adversarial_2018} explored AL-IRL in DPL and use the discriminator to differentiate successful dialogues from unsuccessful ones. Extending this line of research, \citet{takanobu_guided_2019} further combined AL with maximum entropy IRL to learn the policy and reward estimator alternatively.

\subsection{Reward Shaping}
Reward shaping aims to incorporate domain knowledge into RL by introducing an extra reward in addition to the reward provided by the environment \citep{ng_andrew_policy_1999}.
\citet{ferreira_social_2013} learned an extra reward from the social cues of the user. In this work, they mainly consider the sentiment cues from the user-defined manually including the type of dialogue acts, number of slots filled, agenda size etc. While this method doesn't need extra annotated data, the manually defined features are not scalable to other domains. \citet{wang_learning_2020} took advantage of human demonstrations and use a multi-variate Gaussian to pick the most similar state-action pair to complement the main reward. On the whole, these papers highlight the benefit of using a dense reward in DPL. An important difference between inverse reinforcement learning method and reward shaping is that the former learns one single reward function while the latter adds a reward function in addition to the main reward provided by the environment.

\section{Future Direction} \label{future direction}
As the objective of a TDS is to help user to achieve their goal, future researches should aim toward applying TDS in a real-world scenario. There are two main obstacles in our way: the data scarcity problem which can be solved by either domain adaptation or meta policy learning, and lack of robustness in evaluation.

\noindent \textbf{Data Scarcity.} There are many different types of real-world dialogue scenarios such as restaurant booking, weather query, and flight booking etc. It is extremely costly to obtain a large amount of annotated data for different domains. However, the most recent methods presented in this survey often requires a lot of expert demonstrations. As a result, for a TDS to be applicable, we should develop techniques and methods to learn a dialogue policy efficiently and effectively in domains that have scarce data. \textit{Domain Adaptation} and \textit{Meta Policy Learning} are two effective and promising solutions to tackle this problem.

\noindent \textbf{Evaluation Robustness.} It is very important to evaluate the performance of a dialogue policy in assisting humans to complete some tasks. Currently, the most widely used way to evaluate a dialogue policy is to use a user simulator to interact with the dialogue agent and compute some metrics over it. This evaluation method does not correctly reflect how good a dialogue policy can assist a human in completing their task. Below we outline two promising future directions in tackling the data scarcity problem and our insight on a better evaluation method. 

\subsection{Data Scarcity Problem}

\textbf{Domain Adaptation.} Domain adaptation or policy transfer allows us to build a dialogue policy in a target domain that has scarce data provided with a large amount of data in a source domain. \citet{chen_policy_2018} proposed a multi-agent dialogue policy (MADP) that consists of some slot-dependant agents that have shared parameters for every slot. Those shared parameters can be transferred to a new domain for those common slots. In a similar fashion, \citet{ilievski_goal-oriented_2018} matched the state space and action space between the source domain and target domain even if those actions/slots are never used in the source domain. The parameters of the common slots and actions are used in the target domain initially. However, different domains don't necessarily have common actions or consistent dialogue act naming. \citet{mo_cross-domain_2018} proposed a PROMISE model that learns the similarity between the slots and actions of different domains. While these researches focus on domain adaptation between two domains, more works need to be done on adapting to multi-source domains.

\noindent \textbf{Meta Policy Learning.} To further extend the usage of DPL to a real-world scenario, we should consider situations that have an even harsher data resource. In the previous section, we discuss the direction that leverages the abundant data in a source domain. In this section, we consider the meta-learning paradigm that tackles the situation that all domains have scarce data. Recently, \citet{mi_meta-learning_2019} adopted meta-learning in NLG module in the SDS pipeline. Inspired by this work, \citet{xu_meta_2020} proposed Deep Transferable Q-Network (DTQN) that leverages shareable features across domains. They further combine DTQN with Model-Agnostic Meta-Learning (MAML) \cite{finn_model-agnostic_2017} with a dual-replay mechanism to support effective off-policy learning which helps models to adapt to an unseen domain quickly. \citet{zhang_budgeted_2019} extended DDQ by incorporating Budget-Conscious Scheduling to learn from a fixed, small amount of interactions. It uses a decayed poisson process to model the number of interactions allocated to each epoch, where the total number of epochs is predefined. More works are needed to explore efficient learning methods in TDS under the meta-learning paradigm.

\subsection{Evaluation}
In DLP research, \citet{walker_paradise_1997} is the first to present a general framework to evaluate the performance of a dialogue agent. They evaluate a dialogue from two aspects. One is the dialogue cost which measures the cost induced by the dialogue (e.g. number of turns) and another one is task success which evaluates whether the dialogue agent successfully accomplish the task from the user by comparing it with the user goal. In practice, the dialogue policy is often evaluated by having conversations with a simulated user by the metrics such as inform F1, success rate, bleu score \citep{takanobu_is_2020}. The problem is that the simulator doesn't resemble human conversation behaviour well as discussed in Section \ref{sec:environment}. Therefore, there is still a gap between human evaluation and simulated evaluation \citep{takanobu_is_2020}. We believe that much work is needed to provide a universal evaluation framework that should be used for any general TDS. Instead of using metrics that compare the dialogue act with the simulated goal, a universal evaluation framework should emphasize the overall satisfaction of a human user. Such a framework should include but not limited to ways to measure how natural or helpful is the response of the dialogue agent to the user.

\section{Conclusion} \label{conclusion}
In this survey, we introduce the recent advancement of RL approaches applied in DPL of TDS, which focus on tackling the three main challenges. Given the vast amount of works in such areas in recent years, a typology of approaches is needed to identify the main focal research directions in applying RL in DPL. We contribute such a typology that is based on which of the five RL elements the approaches are adapting. As we are moving to apply TDS in real-world scenarios, the scarce data of various dialogue scenarios and the lack of robust evaluation of dialogue agents will be the most prominent obstacles. To this end, three fruitful research directions are suggested to tackle them respectively.

\bibliography{anthology,ref}
\bibliographystyle{acl_natbib}

\appendix

\section{Procedure for Shortlisting Papers} \label{appendix:method}
We use a two-step procedure to shortlist relevant papers for review. In the first step, we use two tools to search relevant papers.  The two tools are 1)  AMiner\footnote{https://www.aminer.cn/} which can provide literature that dates back to 1922 given a topic keyword and 2)  Connected Papers\footnote{https://www.connectedpapers.com/} to provide us with a graph of strongly connected papers given a seed paper. We use Aminer with the keyword "dialogue policy" to search for papers within the recent ten years. Among the returned list of papers, we use each one as a seed paper as input to Connected Papers and further select related papers from the provided graph. Then we go through the papers manually and select those that apply RL methods in DPL of TDS as the preliminary papers. In the second step, we go through the references of the preliminary papers and pick relevant ones. 

\section{Summary of Current Methods} \label{appendix:table}

\begin{sidewaystable*}

    \centering
    \begin{adjustbox}{width=\textwidth}
    \begin{tabular}{*{11}{l}}
        \toprule
       \multirow{2}{*}{Model} &
       \multirow{2}{*}{Dataset} & 
       \multirow{2}{*}{RL algorithm} & 
       \multirow{2}{*}{Experience Replay} & 
       \multicolumn{2}{c}{Simulator}  & 
       \multicolumn{2}{c}{Annotations} & 
       \multicolumn{2}{c}{Expert demo}  &
       \multirow{2}{*}{Reward function}\\
       \cmidrule(lr){5-6}\cmidrule(lr){7-8}\cmidrule(lr){9-10 }
         &&&&Granularity&Methodology&    
         Belief State & Dialogue Act & 
         IL & Supervised Buffer & 
          \\
        \midrule
        TSL \citep{li_temporal_2014} & Calendar & Q-learning & \checkmark & utterance level & rule-based & \checkmark & \checkmark & \checkmark &  & Other\\
        RNN Reward Shaping \citep{su_reward_2015} & CamRes & GP-SARSA &  & dialogue-act level & agenda-based & \checkmark & \checkmark &  &  &  reward shaping \\
        End-to-End RL \citep{zhao_towards_2016} & 20 Question Game & DRQN & \checkmark & utterance level & agenda-based & \checkmark & \checkmark & \checkmark &  & manually defined \\
        Continuous Learning  \citep{su_continuously_2016}  & CamRes & NAC & \checkmark & dialogue-act level & agenda-based & \checkmark & \checkmark & \checkmark &  & manually defined\\
        Two-stage training DQN \citep{fatemi_policy_2016} & DSTC2 & GPSARSA, DA2C, TDA2C, DQN, DDQN & \checkmark & dialogue-act level & agenda-based & \checkmark & \checkmark & \checkmark &   & manually defined \\
        Option Framework \citep{budzianowski-etal-2017-sub} & Pydial & HRL & \checkmark & dialogue-act level & agenda-based & \checkmark & \checkmark & &  & manually defined\\
        BBQN \citep{lipton_bbq-networks_2017} & Amazon Movie-Ticket & DQN & \checkmark & dialogue-act level & agenda-based & \checkmark & \checkmark & \checkmark & & Others \\
        IPLDM \citep{liu_iterative_2017} & DSTC2 & REINFORCE, Multi-Agent&  &  dialogue-act level & multi-agent & \checkmark & \checkmark & \checkmark &  & manually defined \\
        CTCDS \citep{peng_composite_2017} & Frames & HRL & \checkmark & dialogue-act level & agenda-based & \checkmark & \checkmark &  & \checkmark & manually defined \\
        TRACER,eNACER \citep{su_sample-efficient_2017} & CamRes & GPRL,TRPO &\checkmark  & dialogue-act level & rule-based & \checkmark & \checkmark & \checkmark & \checkmark & manually defined \\
        CT \citep{chen_-line_2017} & DSTC2 & DQN & \checkmark & dialogue-act level & agenda-based & \checkmark & \checkmark & & & manually defined \\
        ACER \citep{weisz_sample_2018} & CamRes & Actor-Critic / TRPO / IS & \checkmark & dialogue-act level & agenda-based & \checkmark & \checkmark &  & \checkmark & manually defined\\
        ALDM \citep{liu_adversarial_2018} & DSTC2 & Policy Gradient &  & dialogue-act level & multi-agent & \checkmark & \checkmark & \checkmark &  & AL-IRL \\
        Adversarial A2C \citep{peng_adversarial_2018} & Amazon Movie-Ticket & Actor-Critic & \checkmark & dialogue-act level & agenda-based & \checkmark & \checkmark & \checkmark & \checkmark & AL-IRL\\
        DDQ \citep{peng_deep_2018} &  Amazon Movie-Ticket & Dyna-Q, Actor-Critic & \checkmark & dialogue-act level & world model & \checkmark & \checkmark & \checkmark & \checkmark & manually defined \\
        HER \citep{lu_goal-oriented_2018} &  Amazon Movie-Ticket & DQN & T-HER / S-HER & dialogue-act level & agenda-based & \checkmark & \checkmark &  & & manually defined \\
        FDQN \citep{casanueva2018feudal} & PyDial & Feudal RL & \checkmark & dialogue-act level & agenda-based & \checkmark & \checkmark &  &  & manually defined \\
        Option Framework \citep{kristianto-etal-2018-autonomous} & Pydial & HRL & \checkmark & dialogue-act level & agenda-based & \checkmark & \checkmark & &  &  manually defined \\
        D3Q \citep{su_discriminative_2018} & Amazon Movie-Ticket & Dyna-Q & \checkmark & utterance level & world model & \checkmark & \checkmark & \checkmark & \checkmark &  manually defined \\
        SDN \citep{tang_subgoal_2018} & Frames & HRL & \checkmark & utterance level & agenda-based &  & \checkmark &  &  &  manually defined\\
        Switch-DDQ \citep{wu_switch-based_2018} &  Amazon Movie-Ticket & Dyna-Q & \checkmark & utterance level & world model & \checkmark & \checkmark & \checkmark & \checkmark & manually defined\\
        D3Q \citep{su_discriminative_2018} & Amazon Movie-Ticket & Dyna-Q & \checkmark & utterance level & agenda-based & \checkmark & \checkmark & \checkmark & \checkmark &  manually defined \\
        LaRL \citep{zhao_rethinking_2019} $^{\lozenge}$  &  DealOrNoDeal / MultiWOZ & REINFORCE &  & utterance level & data-driven &  &  &  & \checkmark & manually defined \\
        Meta-DTQN \citep{xu_meta_2020} & MultiWOZ & DQN / Dual Replay & \checkmark & dialogue-act level & agenda-based & \checkmark & \checkmark &  &  & manually defined \\
        WoLF-PHC \citep{papangelis-etal-2019-collaborative} & DSTC2 & WoLF-PHC & & dialogue-act/utterance level & multi-agent &  & \checkmark & & \checkmark & manually defined \\
        BCS-DDQ \citep{zhang_budgeted_2019} & Amazon Movie-Ticket & Dyna-Q & \checkmark & dialogue-act level & world model & \checkmark & \checkmark &  & \checkmark & manually defined \\
        GDPL \citep{takanobu_guided_2019} & MultiWOZ & PPO &  & dialogue-act level & agenda-based & \checkmark & \checkmark &  & \checkmark & AL-IRL \\
        LHUA \citep{cao_adaptive_2020} & Amazon Movie-Ticket & DQN & T-HER / S-HER & dialogue-act level & agenda-based & \checkmark & \checkmark & & & manually defined\\
        Act-VRNN \citep{huang_semi-supervised_2020}& MultiWOZ & ELBO & \checkmark & dialogue-act level & agenda-based & \checkmark & \checkmark & & \checkmark & Others\\
        OPPA \citep{zhang_learning_2020} & MultiWOZ &  DQN & \checkmark & dialogue-act level & agenda-based &   & \checkmark & \checkmark &  &  manually defined \\ 
        GDPL w/o AL\citep{li_guided_2020} & MultiWOZ & PPO, DQN & \checkmark & dialogue-act level & rule-based & \checkmark & \checkmark &  &  & AL-IRL\\
        MADPL \citep{takanobu_multi-agent_2020}  & MultiWOZ & Actor-Critic, Multi-Agent & \checkmark & dialogue-act level & multi-agent & \checkmark & \checkmark & \checkmark &  & manually defined\\     
       	DQfD \citep{gordon-hall_show_2020} & MultiWOZ & DQN &\checkmark  & dialogue-act level & agenda-based & \checkmark & \checkmark &  & \checkmark &  manually defined\\
       	RoFL \citep{gordon-hall_learning_2020} & MultiWOZ & DQN &\checkmark  & dialogue-act level & agenda-based & \checkmark, \footnotemark & \checkmark , & \checkmark & \checkmark &  manually defined\\
   
        \bottomrule 
    \end{tabular}
    \end{adjustbox}
    \caption{An overview of the configurations of recent works on DPL with RL approach. }
    \label{tab:summary_table}
\end{sidewaystable*}
\footnotetext{This paper proposed three models that work on data with belief state and dialogue act annotations, dialogue act annotations only and without any annotations respectively.}

\end{document}